\documentclass[cameraready]{Interspeech}

\usepackage{CJKutf8}

\title{Human-like cross-language generalisation in deep neural speaker embeddings and its acoustic foundations}

\author[affiliation={1}, orcid=0009-0009-9207-4149, correspondingauthor, equalcontribution]{Tianze}{Xu}
\author[affiliation={1}, orcid=0009-0004-3456-6923, equalcontribution]{Xiyang}{Li}
\author[affiliation={2}, orcid=0000-0002-5171-9774]{Xiaoming}{Jiang}
\author[affiliation={1}, orcid=0000-0002-8494-6025]{Volker}{Dellwo}

\address{
    $^1$ Department of Computational Linguistics, University of Zurich, Switzerland \\
    $^2$ Institute of Language Sciences, Shanghai International Studies University, China
}

\email{tianze.xu@uzh.ch, xiyang.li2@uzh.ch, xiaoming.jiang@shisu.edu.cn, volker.dellwo@uzh.ch}

\keywords{speaker embeddings, cross-language generalisation, voice similarity, human--machine alignment}

\usepackage{comment}

\begin{document}

\maketitle

\begin{abstract}
    Generalisability is a central question for deep neural networks, including the cross-language generalisation of speaker embeddings. This study examined whether deep neural speaker embeddings generalise across languages and how such generalisation relates to human perception and acoustic structure. Using Cantonese–English bilingual speech presented in within- and cross-language conditions, we derived similarity from 18 state-of-the-art models and compared it with human ratings and acoustic features using linear mixed models and representational similarity analysis. Embeddings exhibited cross-language generalisation that broadly paralleled human behaviour, reflecting accurate identity discrimination and sensitivity to language switching. This generalisation relied on largely shared acoustic cues with language-sensitive weighting, mirroring human patterns. Together, these findings provide converging evidence for human–machine alignment in cross-language voice similarity processing.\footnote{This manuscript has been accepted for presentation at Interspeech 2026. Code is available at \url{https://github.com/xiyangg12/wespeaker}. The archived release is available on Zenodo at \url{https://doi.org/10.5281/zenodo.20606535}.}
\end{abstract}

\section{Introduction}

In speech and language processing, a pressing contemporary goal is to understand whether and how neural network models can generalise---namely, transfer representations, knowledge, and strategies acquired from prior experience to novel inputs \cite{Hupkes2023,Lake2023}. One important case concerns cross-language generalisation (e.g., from English to Cantonese). In speaker modelling (typically via speaker embeddings; see \cite{Wang2024b} for a review), this issue is particularly relevant because languages differ systematically in their phonetic and phonological structure. Such differences can modulate acoustic realisations of voice, potentially challenging downstream applications including cross-lingual speech synthesis \cite{Zhang2024b,Zhou2023} and automatic speaker recognition \cite{Liu2023}.

By contrast, humans are widely recognised as exhibiting a remarkable capacity for generalisation \cite{Lake2023,Taylor2021}. For example, Xu and Jiang \cite{Xu2026} showed that listeners generalise across languages when judging voice similarity. English-speaking participants with and without Cantonese proficiency rated the voice similarity of speech pairs presented in Cantonese-only, English-only, or mixed-language conditions. Same-speaker pairs were consistently judged more similar than different-speaker pairs. Mixed-language pairs were rated as more dissimilar than Cantonese- or English-only pairs, particularly for same-speaker comparisons, suggesting that language switching disrupts vocal cue integration. Crucially, listeners’ language background did not significantly affect ratings, indicating that English monolingual listeners exhibited robust cross-language generalisation. Acoustic--perceptual correlations further identified pitch, vocal tract configuration, and harmonic/noise spectral balance as key cues whose weighting varied by language context.

Indeed, similarity judgement is a fundamental cognitive operation \cite{Roads2024a}, providing a principled testbed for human–machine alignment. Unlike tasks that rely on long-term memory for humans (e.g., recognition \cite{Perrachione2011}), similarity evaluation can be modelled more directly in both humans and machines. Demonstrating this, Liu et al. \cite{Liu2024} compared perceptual voice similarity with similarity derived from embeddings from seven deep neural network models using English speech. They found no significant correlations between perceptual and machine similarity under fine-grained control, whereas machine similarity correlated with acoustic similarity computed from an aggregate of 24 acoustic variables. Importantly, this study did not examine cross-linguistic generalisation. Other work has done so, but primarily using earlier models based on Mel-frequency cepstral coefficients \cite{Gerlach2020, Gerlach2023}. For instance, Gerlach et al. \cite{Gerlach2020} reported positive correlations between perceptual and machine similarity for English speech, observed in both English and German listeners, with stronger effects for English listeners.

Despite substantial engineering efforts to improve the cross-language robustness of speaker embeddings for downstream applications (e.g., \cite{Zhang2024b,Zhou2023,Liu2023}), it remains underexamined whether and how deep neural speaker embeddings generalise across languages, particularly in relation to human perception. The present study addresses this gap by directly comparing machine, perceptual, and acoustic similarity structures across within- and cross-language conditions using Cantonese--English bilingual speech. Cantonese and English differ markedly in phonological structure, offering a stringent test of cross-language generalisation while enabling direct comparison with existing perceptual data \cite{Xu2026}. We investigated what drives state-of-the-art speaker embeddings' capture of voice similarity using linear mixed models (LMMs). Applying representational similarity analysis (RSA), we examined how speaker embeddings relate to human listeners' representations of speaker identity and the fine-grained acoustic features underlying these representations. To ensure generalisability, the models analysed spanned a wide range of architectures, parameter scales, and pretraining corpora.
Crucially, language-background effects were tested by comparing pretrained and Cantonese fine-tuned embeddings, providing a principled assessment of human--machine alignment in voice processing.

\section{Methods}
Speech and perceptual data were drawn from \cite{Xu2026}; the present study further derived acoustic measures and speaker embeddings from the speech recordings for subsequent analyses.

\subsection{Speech data}
Speech recordings were sourced from the SpiCE corpus \cite{Johnson2020b}, which comprises recordings from early Cantonese--English bilinguals. Xu and Jiang \cite{Xu2026} included 10 female speakers (aged 19--32 years, $M=22.60$, $SD=3.89$) who were highly proficient in Hong Kong Cantonese and North American English. Two semantically comparable festive greeting recordings were selected in each language: Cantonese (duration 1.17--1.43 s, $M=1.34$, $SD=0.12$): “\begin{CJK*}{UTF8}{bsmi}新年快樂\end{CJK*}” (\textit{Happy New Year}) and “\begin{CJK*}{UTF8}{bsmi}恭喜發財\end{CJK*}” (\textit{Congratulations on happiness and prosperity}); English (duration 1.18--1.87 s, $M=1.45$, $SD=0.17$): \textit{Have a spooky Halloween} and \textit{Relax on your holiday break}.

\subsection{Perceptual data}

Forty participants (aged 18--32 years, $M = 22.25$, $SD = 3.75$; 20 native Cantonese--English bilinguals and 20 native English speakers with no Cantonese knowledge) completed a voice similarity rating task. On each trial, participants listened to two speech samples and rated voice similarity on a 9-point scale (1 = \textit{very similar}, 9 = \textit{very different}). After 10 practice trials, 220 main trials were presented in four randomised blocks. The session was self-paced and lasted about 45 min.

Two utterances in Cantonese and two in English produced by each of the 10 speakers yielded four 10~×~10 speaker--utterance matrices (excluding identical-utterance pairs). Assuming symmetry in similarity judgments, only the lower triangle of each matrix (including the same-speaker pairs on the diagonals) was sampled, resulting in 220 trials in total. Trials varied along two factors: speaker identity (40 same-speaker vs. 180 different-speaker pairs) and stimulus language (55 Cantonese-only, 55 English-only, and 110 mixed-language).

\subsection{Acoustic measurements}
We analysed psychoacoustic features in five domains \cite{Xu2026,Kreiman2021}:
\begin{itemize}
\item \textbf{Temporal}: speech rate (SR; number of syllables per second).
\item \textbf{Pitch:} $f_0$ (reflecting the rate of vocal fold vibration).
\item \textbf{Formant structure}: $F_1$--$F_4$ and formant dispersion (FD; an index of vocal tract length) \cite{Fitch1997}.
\item \textbf{Harmonic source}: relative amplitudes of selected harmonics (i.e., $H_1^{\ast}-H_2^{\ast}$, $H_2^{\ast}-H_4^{\ast}$, $H_4^{\ast}-H_{2\text{kHz}}^{\ast}$, and $H_{2\text{kHz}}^{\ast}-H_{5\text{kHz}}$), encoding phonatory characteristics \cite{Kreiman2024,Garellek2013a}.
\item \textbf{Inharmonic source}: root-mean-square (RMS) energy (reflecting signal amplitude), cepstral peak prominence (CPP) \cite{Fraile2014}, subharmonics-to-harmonics ratio (SHR) \cite{Sun2002}, and harmonics-to-noise ratio (HNR) \cite{Perrachione2019a}.
\end{itemize}
All measures except SR and FD were extracted every 5~ms using VoiceSauce \cite{Shue2011} from voiced segments only. FD was computed as the mean successive difference across $F_1$--$F_4$, and SR from utterance duration and syllable count. For each utterance of each speaker, we calculated means and coefficients of variation (CVs) for all measures except SR and harmonic spectral shape measures. SR was represented by a single mean value. Because the mean values of harmonic measures can be zero, we computed SD instead of CV (differing from \cite{Xu2026}) and mean. In total, 29 acoustic variables were obtained per speaker--utterance.

\subsection{Speaker embedding extraction}

Speaker embeddings were extracted using the Wespeaker (Version~1.2.0) toolkit \cite{Wang2024a} in Python. To ensure generalisability of the findings, we employed 18 pretrained models spanning a wide range of architectures, parameter scales, and training data:
\begin{itemize}
\item \textbf{Seven residual network (ResNet) models} \cite{Zeinali2019}: ResNet34-VX, ResNet34LM-VX, ResNet152LM-VX, ResNet221LM-VX, ResNet293LM-VX, ResNet34-CN, and ResNet34LM-CN. ResNet is a deep convolutional architecture with skip connections that enable very deep training \cite{He2016}.
\item \textbf{Four simple attention module (SimAM) ResNet models} \cite{Qin2022}: SimAMResNet34-VB, SimAMResNet34-VBVX, SimAMResNet100-VB, and SimAMResNet100-VBVX. SimAM is a lightweight, parameter-free attention mechanism that enhances feature discrimination \cite{Yang2021}.
\item \textbf{Five emphasized channel attention, propagation, and aggregation (ECAPA) models} \cite{Desplanques2020}: ECAPA1024-VX, ECAPA1024LM-VX, ECAPA512-VX, ECAPA512LM-VX, and ECAPA512DINO-VX. ECAPA enhances channel attention and multi-scale feature aggregation within time-delay neural network (TDNN) \cite{Snyder2018} architectures to improve speaker discrimination \cite{Desplanques2020}. \textit{DINO} denotes self-\textit{di}stillation of knowledge with \textit{no} labels during pretraining \cite{Caron2021}.
\item \textbf{Two context-aware masking (CAM++) models} \cite{Wang2023a}: CAM++-VX and CAM++LM-VX. CAM++ employs context-aware masking and multi-granularity pooling to enhance contextual feature modelling \cite{Wang2023a}.
\end{itemize}
Numbers denote network depth. Models marked \textit{LM} were trained with large-margin softmax objectives \cite{Thienpondt2021}. The suffixes \textit{-VX}, \textit{-CN}, and \textit{-VB} indicate pretraining on the VoxCeleb \cite{Nagrani2020}, CN-Celeb \cite{Li2022}, and VoxBlink2 \cite{Lin2024} corpora, respectively. \textit{-VBVX} denotes VoxBlink2 pretraining followed by VoxCeleb2 fine-tuning. VoxCeleb and VoxCeleb2 primarily contain English-speaking celebrity speech, whereas CN-Celeb comprises speech from Chinese celebrities (largely Mandarin sources, with unspecified dialect distribution). Thus, despite being state-of-the-art models pretrained on large-scale datasets, their validity for Cantonese speakers remains uncertain.

To expose the models to Cantonese, we fine-tuned them on the ASR-SCCantDuSC \cite{MagicData2024a} ($\sim$4 hours, 10 speakers) and the ASR-SCCantCabSC \cite{MagicData2024b} ($\sim$5 hours, 10 speakers) corpora. 
Using the Wespeaker toolkit, pretrained weights were updated on the new data under the original training objectives, allowing adaptation to Cantonese characteristics while preserving general representations from large-scale pretraining.

Speaker embeddings were extracted using the 18 pretrained models and their Cantonese fine-tuned counterparts. Each utterance from each speaker was passed through each model to obtain high-dimensional embeddings.

\subsection{Data analyses}
To investigate factors influencing machine representations of speaker identity, we fitted an LMM predicting cosine similarity \cite{Wang2024a} between embeddings in each speech pair. Fixed effects were speaker identity (same vs. different), stimulus language (Cantonese-only, English-only, mixed), fine-tuning status (original pretrained vs. Cantonese fine-tuned), their full interactions, and model architecture (ResNet, SimAM, ECAPA, CAM++). Random intercepts were specified for model and speech pair, with a random slope for speaker identity by model retained after stepwise reduction \cite{Meteyard2020}. 

RSAs \cite{Kriegeskorte2008} were then used to examine how machine representations relate to perceptual and acoustic structures. We first constructed perceptual, acoustic, and machine representational dissimilarity matrices (RDMs). Perceptual RDMs were drawn from \cite{Xu2026} and comprised 220 speech pairs per participant, yielding 40 RDMs (20 Cantonese--English bilinguals, 20 English monolinguals). Acoustic RDMs (29 in total) were computed as absolute differences between paired speech samples for each variable. Machine RDMs were defined as $1-$cosine similarity between embeddings for each speech pair, resulting in 36 RDMs (18 original and 18 fine-tuned models). Each machine RDM was correlated with perceptual or acoustic RDMs using Spearman’s rank correlation, separately by stimulus language, listener language background, and fine-tuning status.

Statistical significance was assessed using one-sided Wilcoxon signed-rank tests \cite{Wilcoxon1992} with false discovery rate (FDR) correction ($\alpha=.05$). Correlations from the machine--perceptual RSA were analysed using an LMM, with fixed effects of stimulus language, listener language background (Cantonese--English vs. English), fine-tuning status, their full interactions, and architecture. Random intercepts were specified for model, with no random slopes retained following stepwise model reduction. Correlations from the machine--acoustic RSA were compared across fine-tuning status using paired Wilcoxon signed-rank tests.

\section{Results}

\subsection{LMM on machine-derived similarity}

\begin{figure}[htbp]
  \centering
  \includegraphics[width=\linewidth]{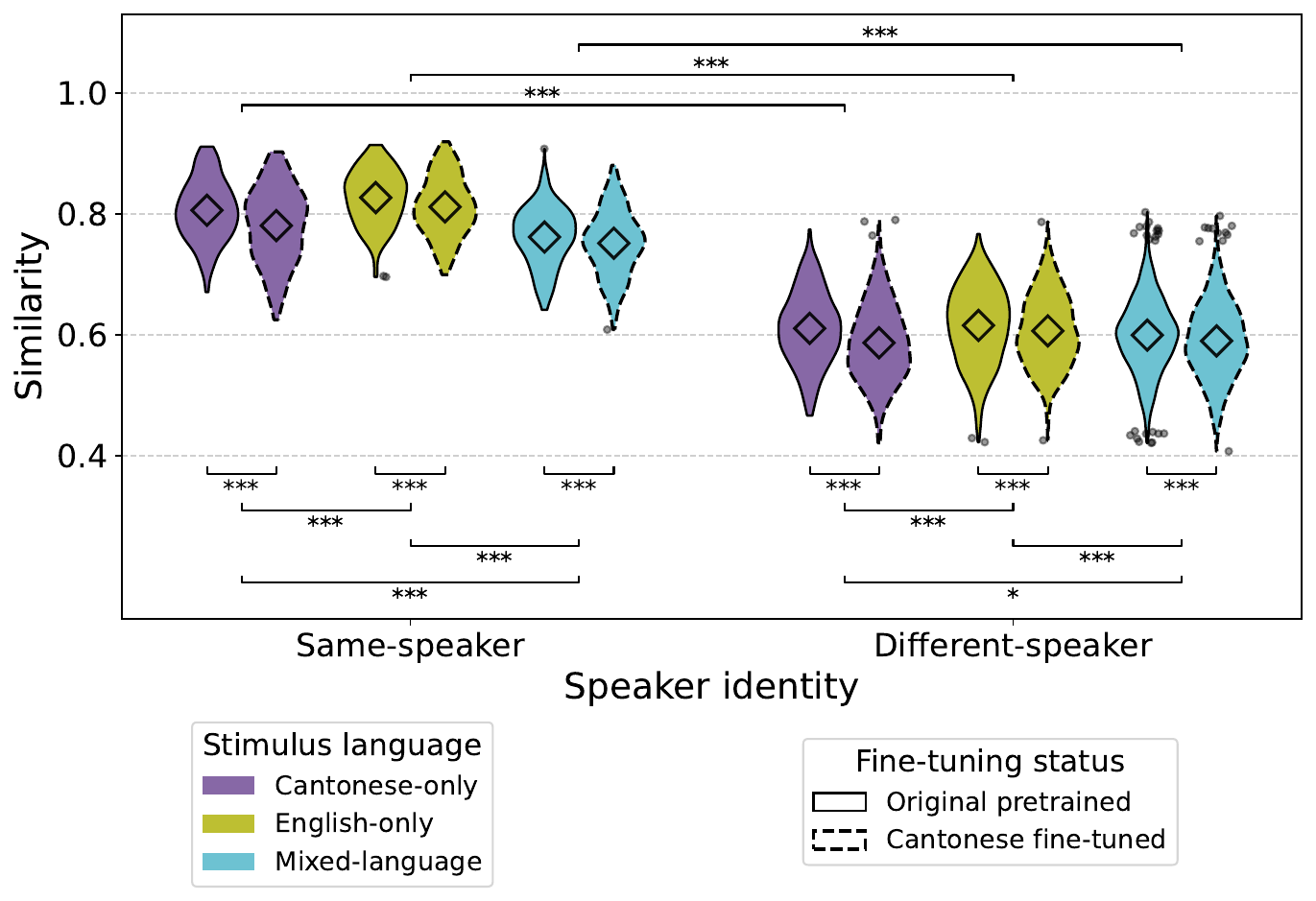}
  \caption{Machine-derived cosine similarity score distributions with means (diamonds) and outlier points. \textit{n.s.}, $p > .050$; \textit{*}, $p < .050$; \textit{**}, $p < .010$; \textit{***}, $p < .001$.}
  \label{fig:machine}
\end{figure}

Machine similarity scores ranged from 0.41 to 0.92 ($M = 0.63$, $SD = 0.09$; Figure~\ref{fig:machine}). The LMM revealed significant main effects of speaker identity ($\chi^2(1)=671.39$, $p<.001$), stimulus language ($\chi^2(2)=175.15$, $p<.001$), fine-tuning status ($\chi^2(1)=17.30$, $p<.001$), and architecture ($\chi^2(3)=94.70$, $p<.001$). Among the interactions, only speaker identity $\times$ stimulus language was significant ($\chi^2(2)=81.49$, $p<.001$). Neither speaker identity $\times$ fine-tuning status ($\chi^2(1)=0.05$, $p=.827$) nor stimulus language $\times$ fine-tuning status ($\chi^2(2)=4.15$, $p=.126$) reached significance. The three-way interaction was also not significant ($\chi^2(2)=0.53$, $p=.765$).


For the main effect of speaker identity, machine similarity was higher for same-speaker ($M = 0.78$) than for different-speaker ($M = 0.60$) pairs. For stimulus language, English-only pairs showed the highest mean similarity ($M = 0.65$), followed by Cantonese-only ($M = 0.63$) and mixed-language ($M = 0.62$) pairs. For fine-tuning status, original pretrained models yielded slightly higher similarity ($M = 0.64$) than Cantonese fine-tuned models ($M = 0.63$). Among architectures, SimAM achieved the highest similarity ($M = 0.65$), followed by ResNet, ECAPA, and CAM++ (all $M = 0.63$). For speaker identity $\times$ stimulus language, post hoc pairwise comparisons with Holm correction revealed that the speaker identity effect (same--difference contrast) was attenuated for mixed-language pairs ($\Delta M = 0.16$, $z = 25.61$, $p < .001$) compared with Cantonese-only ($\Delta M = 0.19$, $z = 28.82$, $p < .001$) or English-only pairs ($\Delta M = 0.21$, $z = 30.85$, $p < .001$).


\subsection{RSA between machine and perceptual representations}
Machine and perceptual RDMs showed moderately strong correlations overall (range .34–.57, $M=.44$, $SD=0.05$, all $p<.001$). The LMM revealed significant main effects of stimulus language ($\chi^2(2)=118.95$, $p<.001$) and listener language background ($\chi^2(1)=39.18$, $p<.001$), but not fine-tuning status ($\chi^2(1)=1.86$, $p=.173$) or architecture ($\chi^2(3)=6.43$, $p=.093$). Among the interactions, only stimulus language $\times$ listener language background was significant ($\chi^2(2)=31.20$, $p<.001$); neither stimulus language $\times$ fine-tuning status ($\chi^2(2)=0.15$, $p=.928$), listener language background $\times$ fine-tuning status ($\chi^2(1)=1.54$, $p=.215$), nor the three-way interaction ($\chi^2(2)=0.85$, $p=.654$) reached significance. Pairwise comparisons for stimulus language $\times$ listener language background showed that correlations were higher for Cantonese--English bilingual listeners than English monolinguals in the Cantonese-only condition ($\Delta M = 0.053$, $z = 7.61$, $p < .001$), slightly lower in the English-only condition ($\Delta M = -0.015$, $z = -2.23$, $p = .027$), and did not differ reliably in the mixed-language condition ($\Delta M = 0.011$,  $z = 1.62$, $p = .107$). However, no effects involving fine-tuning status were observed.

\subsection{RSA between machine and acoustic representations}
\begin{figure*}[t]
  \centering
  \includegraphics[width=0.886\linewidth]{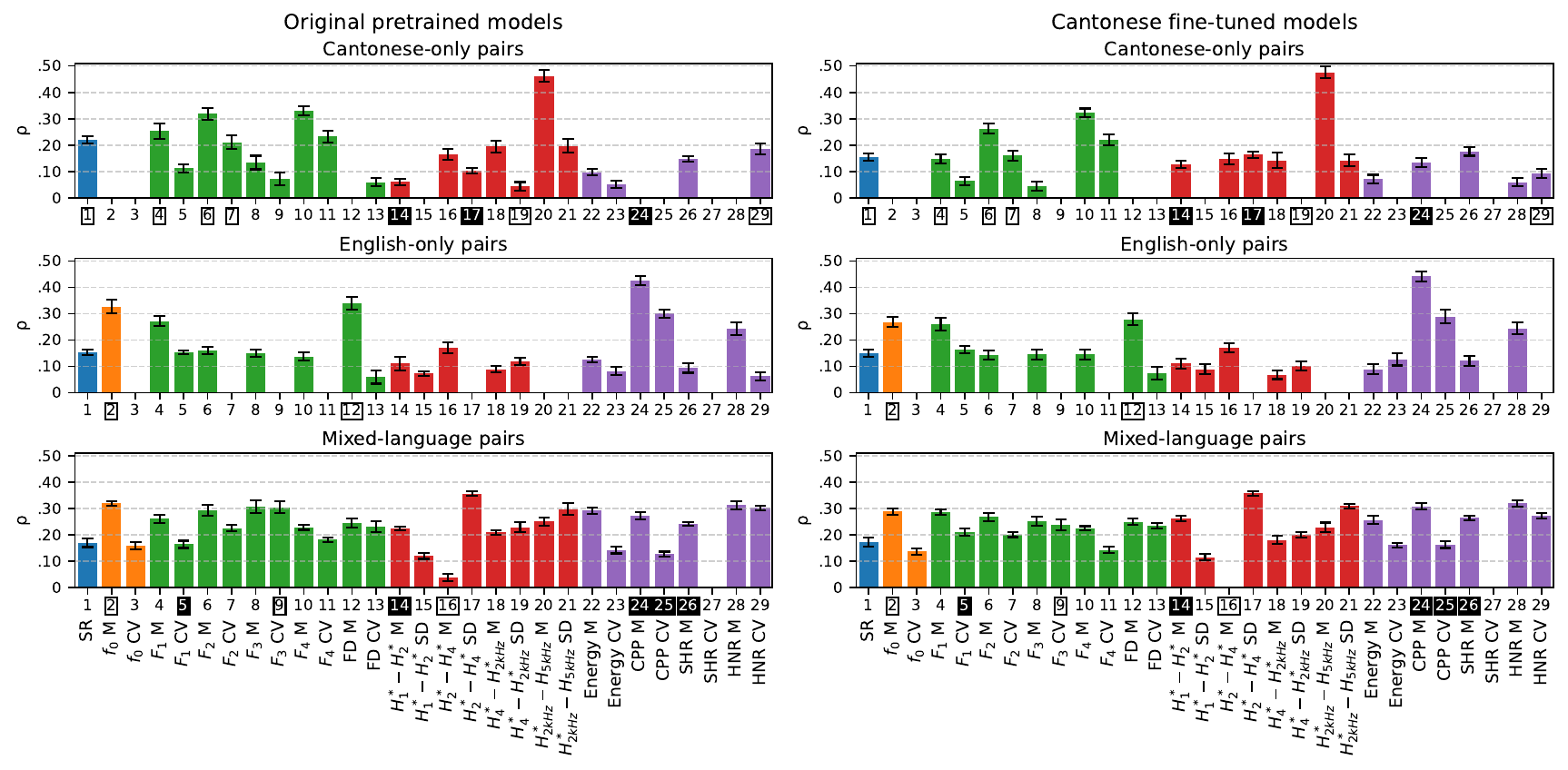}
  \caption{Spearman’s rank correlation coefficients ($\rho$) between machine and acoustic RDMs, averaged across models within each fine-tuning status, are shown for each stimulus language condition.  Acoustic variables are numbered on the x-axis, with squares marking significant changes after fine-tuning (white: decrease, black: increase). Bars are coloured when correlations are significant ($\alpha=.05$, FDR-corrected), with colours indicating acoustic feature categories. Error bars indicate the standard error of the mean (SEM).}
  \label{fig:acoustics}
\end{figure*}

Machine--acoustic RSA identified acoustic features significantly associated with machine-derived similarity (Figure~\ref{fig:acoustics}). Across language and fine-tuning conditions, the strongest correlates included SR and the means of $F_1$–$F_4$ and SHR. Some language-specific patterns emerged. For Cantonese-only pairs, the strongest correlates were the means of $F_2$ ($\rho = .29$), $F_4$ ($\rho = .33$), and $H_{2\text{kHz}}^{\ast}{-}H_{5\text{kHz}}$ ($\rho = .47$), averaged across original and fine-tuned models. For English-only pairs, the strongest ones were the means of $f_0$ ($\rho = .30$), FD ($\rho = .31$), and CPP ($\rho = .43$) and CPP CV ($\rho = .29$). Mixed-language pairs showed a more balanced pattern. Furthermore, correlations were largely stable before and after fine-tuning, with only a few significant changes (tested only when the original or fine-tuned correlation was already significant): nine for Cantonese-only pairs (e.g., $F_1$ mean, CPP mean, HNR CV), two for English-only pairs ($f_0$ mean and FD mean), and eight for mixed pairs (e.g., $F_1$ CV, $F_3$ CV, SHR mean). However, the magnitude of these differences was small (all $|\Delta \rho| < 0.12$).


\section{Discussion}

This study aimed to examine whether deep neural speaker embeddings generalise across languages and how it relates to human perception and acoustic structure. Using bilingual speech samples paired in Cantonese-only, English-only, and mixed-language conditions, we derived similarity from embeddings extracted from 18 state-of-the-art models spanning diverse architectures, parameter scales, and training corpora. An LMM was fitted to identify predictors of machine similarity, and RSAs compared machine similarity with perceptual similarity from listeners with different language backgrounds and with acoustic similarity computed from each of the 29 validated variables.

The LMM on machine similarity revealed three key patterns. First, same-speaker pairs consistently received higher similarity than different-speaker pairs, indicating accurate identity discrimination. Second, this same--different contrast was attenuated in the mixed-language condition compared with Cantonese-only or English-only, suggesting that language switching disrupts identity cue integration. Third, Cantonese fine-tuning did not improve similarity estimates for Cantonese pairs: original pretrained models yielded higher similarity scores than fine-tuned models across all conditions (Figure~\ref{fig:machine}). These findings closely parallel human behaviour \cite{Xu2026,Perrachione2019a}; for instance, on the same speech materials, human listeners showed no effect of language background on similarity ratings, consistent with our observation here \cite{Xu2026}.

Machine--perceptual RSA revealed moderately strong correlations across stimulus language, listener language background, and fine-tuning status conditions, suggesting that speaker embeddings exhibit cross-language generalisation broadly comparable to human performance. However, this generalisation is not perfect: both humans and models occasionally make errors (e.g., low similarity ratings for same-speaker pairs).

Machine--acoustic RSA showed that machine similarity was most strongly associated with SR, the means of $F_1$–$F_4$ and SHR. Language-specific patterns emerged for some features, with stronger correlations for $F_2$, $F_4$, and $H_{2\text{kHz}}^{\ast}{-}H_{5\text{kHz}}$ in Cantonese-only pairs, and for $f_0$, FD, CPP, and CPP CV in English-only pairs, while mixed-language pairs exhibited a more balanced profile. This result mirrors findings in human listeners \cite{Xu2026} and production-based evidence \cite{Johnson2023,Kreiman2025}. Correlations remained largely stable after fine-tuning, though more variables changed for Cantonese-only and mixed-language pairs than for English-only pairs, likely reflecting exposure effects on Cantonese-specific properties. Together, these results suggest that both machines and humans rely on partially language-sensitive but largely shared acoustic cues, supporting cross-language generalisation in voice similarity processing.

Importantly, the absence of a fine-tuning benefit is unlikely due to the relatively small size of the Cantonese datasets. If additional exposure were effective, fine-tuned models should have shown higher similarity for same-speaker Cantonese pairs and lower similarity for different-speaker pairs relative to original models. Instead, only the latter pattern emerged (and uniformly in other conditions), while same-speaker similarity decreased. Insufficient data would more plausibly yield no change than a reversal. Nevertheless, future work could examine large-scale Cantonese fine-tuning \cite{Li2026}.
It would also be informative to use longer speech samples from more speakers for both machines and humans \cite{Liu2024}, while balancing listener fatigue. More broadly, our finding that speaker embeddings exhibit genuine cross-language generalisation has important implications for the transfer of speaker models to low-resource languages \cite{Vajrobol2025}.

\section{Conclusion}

This study showed that deep neural speaker embeddings exhibit robust (though not perfect) cross-language generalisation that broadly parallels human voice similarity perception. Across models and fine-tuning conditions, machine similarity reflected accurate identity discrimination, sensitivity to language switching, and stable alignment with perceptual and acoustic structures. Embeddings relied on largely shared acoustic cues with language-sensitive weighting, mirroring human patterns. Together, these findings suggest that contemporary speaker embeddings capture acoustically grounded and cognitively relevant aspects of voice representation, providing evidence for human--machine alignment in cross-language similarity processing and informing transfer to low-resource languages.

\section{Acknowledgments}

\ifcameraready
    This research was partly supported by the Marie Skłodowska-Curie Actions Doctoral Networks under the European Union’s Horizon Europe programme (Voice Communication Sciences; Grant No. 101168998), funded by the Swiss State Secretariat for Education, Research and Innovation (Grant No. 24.00438). The authors thank Eleanor Chodroff for her helpful comments.
\else
\fi


\section{Generative AI Use Disclosure}

ChatGPT was used for language editing and stylistic refinement of the manuscript (e.g., improving clarity and grammar). All authors take full responsibility for the content of the paper and consent to its submission.


\bibliographystyle{IEEEtran}
\bibliography{mybib}

\end{document}